\RequirePackage{fix-cm}
\documentclass[twocolumn]{svjour3}          
\smartqed  

\usepackage{lineno,hyperref}
\usepackage{amssymb}
\usepackage{graphicx}
\usepackage[ruled,vlined,linesnumbered]{algorithm2e}
\usepackage{float}
\usepackage{color}
\usepackage{soul}
\usepackage{amsmath}
\usepackage{caption}
\usepackage{makecell}
\usepackage{enumitem} 
\usepackage{makecell}
\usepackage{subcaption}
\usepackage{multirow,siunitx}
\begin{document}
\title{BehavDT: A Behavioral Decision Tree Learning to Build User-Centric Context-Aware Predictive Model}

\titlerunning{A Behavioral Decision Tree Learning}        

\author{Iqbal H. Sarker$^{1,2,*}$, Alan Colman$^{1}$, Jun Han$^{1}$, Asif Irshad Khan$^{3}$, Yoosef B. Abushark$^{3}$ and Khaled Salah$^{4}$}

\institute{$^1$ Swinburne University of Technology, Melbourne, VIC-3122, Australia. \\ $^2$ Chittagong University of Engineering and Technology, Chittagong, Bangladesh. \\ $^3$ King Abdulaziz University, Jeddah, Saudi Arabia. \\ $^4$ Khalifa University, UAE. \\  $*$Corresponding Email: msarker@swin.edu.au}


\date{Received: date / Accepted: date}

\maketitle

\newcommand\blfootnote[1]{%
	\begingroup
	\renewcommand\thefootnote{}\footnote{#1}%
	\addtocounter{footnote}{-1}%
	\endgroup
}

\makeatletter
\def\footnoterule{\relax%
	\kern 0pt
	\hbox to \columnwidth{\hfill\vrule width .95\linewidth height 0.6pt\hfill}
	\kern .1pt}
\makeatother

\blfootnote{\textit{[A print version published in the Journal: Mobile Networks and Applications, Springer.]}}

\begin{abstract}
This paper formulates the problem of building a \textit{context-aware predictive model} based on user diverse behavioral activities with smartphones. In the area of machine learning and data science, a tree-like model as that of decision tree is considered as one of the most popular classification techniques, which can be used to build a data-driven predictive model. The traditional decision tree model typically creates a number of leaf nodes as decision nodes that represent context-specific rigid decisions. However, in many practical scenarios within the context-aware environment, the \textit{generalized} outcomes could play an important role to effectively capture user behavior. In this paper, we propose a \textit{behavioral decision tree}, ``BehavDT'' context-aware model that takes into account user \textit{behavior-oriented generalization} according to individual preference level. The BehavDT model outputs not only the generalized decisions but also the context-specific decisions in relevant exceptional cases. The effectiveness of our BehavDT model is studied by conducting experiments on individual user real smartphone datasets. Our experimental results show that the proposed BehavDT context-aware model is more effective when compared with the traditional machine learning approaches, in predicting user diverse behaviors considering multi-dimensional contexts.

\end{abstract}

\section{Introduction}
Nowadays, machine learning based predictive analytics and context-aware computing have become active research areas, particularly for building user-centric intelligent and adaptive systems in the domain of Internet of Things (IoT) \cite{sarker2019context}. Smartphones are considered as one of the most important IoT devices due to the popularity of smartphone-enabling technologies including sensors, ubiquitous connectivity, context-awareness etc. in the environment of IoT \cite{el2017smartphone}. These smartphones are frequently used by the users worldwide, in their various daily life activities such as voice communication, apps usage, social networking, healthcare services, traffic monitoring, tourist guide, online shopping etc. These devices can record various contextual information and corresponding users' activities with their phones through the device logs \cite{sarker2018mobile}. Predicting user context-aware behavioral activities utilizing phone log datasets can be used to build various intelligent assistive systems, e.g., context-aware intelligent recommendation systems \cite{sarker2018BehavMiner} \cite{sarker2019AppsPred} \cite{zhu2014mining}. Throughout this paper, we consider users' phone call assistive services by effectively predicting their diverse behavioral activities in relevant contexts, as an example of our context-aware model. Users' such behavioral activities with their phones are not static in the real world, may vary in different contexts, such as ``when'' that represents temporal context, ``where'' that represents spatial context, or ``why'' that might represent social context etc. In this paper, we aim to design a \textit{behavioral decision tree} machine learning classification approach for the purpose of building an effective context-aware predictive model based on these multi-dimensional contexts.

Researchers have proposed several popular machine learning classification algorithms such as ZeroR, Naive Bayes, Decision Tree, Support Vector Machine, K-Nearest Neighbors, Logistic Regression etc. with the ability of building a prediction model \cite{han2011data}. Among the classification approaches, a tree-like model, i.e., decision tree model is considered as one of the most popular and well-known approach in the area of data-driven context-aware computing. In particular, a number of researchers \cite{zulkernain2010mobile} \cite{hong2009context} \cite{lee2007deploying}  \cite{sarker2019machine} \cite{sarker2017effective} have used decision tree classification technique to model mobile users' behavior for various purposes. However, in many user-centric context-aware cases, decision tree classification rules are less performing and low reliable while making corresponding behavioral decisions of a smartphone user \cite{sarker2019context}. According to \cite{freitas2000understanding}, decision trees cannot ensure that a discovered classification rule will have a high predictive accuracy due to over-fitting problem and inductive bias. Such issues may arise because of \textit{lacking generalization} while making decisions in a user-centric context-aware model. Moreover, the traditional decision tree model does not provide the flexibility to set users' preferences that may vary from user-to-user according to their behavioral consistency, leading to \textit{rigid decision making}. Thus, the research question addressing in this paper is - \textit{How to minimize these issues and build an effective user-centric context-aware predictive model?}

In order to answer this research question, in this paper, we present a \textit{behavioral decision tree (BehavDT)} context-aware model that minimizes the above mentioned issues and makes more accurate decisions for unseen test cases. This paper significantly revises our earlier paper \cite{sarker2017approach}, particularly, in terms of designing generalized tree and experimental analysis. In our BehavDT model, once the dataset is ready to process, we construct a behavioral decision tree rather than the traditional decision tree \cite{quinlan1993}. We calculate the entropy and information gain to select the best contextual feature, and perform \textit{behavior-oriented generalization} while creating the nodes. The generalization is taken into account for a particular confidence level preferred by the users, which may vary from user-to-user in the real world. While generating nodes, the corresponding confidence value is calculated by determining the ratio of the \textit{dominant behavior} for the associated context values. In our BehavDT model, the number of produced nodes is not static, may differ depending on the preferred confidence level, because of performing \textit{behavior-oriented generalization}. Once the complete tree has been built, both the interior nodes and leaf nodes are used to make the context-aware decisions. As a result, our BehavDT context-aware model is able to capture not only the general behavior of an individual user but also the specific exceptions in relevant multi-dimensional contexts, and improves the prediction accuracy for unseen test cases.

The contributions of this work can be summarized as follows.

\begin{itemize}
	\item We propose a \textit{behavior-oriented generalization} approach while designing a decision tree. 
	
	\item We present a \textit{behavioral decision tree (BehavDT)} context-aware model for effectively predicting user-centric context-aware cases considering relevant multi-dimensional contexts.
	
	\item We have conducted \textit{experiments} on the real mobile phone datasets to evaluate our BehavDT model on unseen context-aware test cases.
\end{itemize}

The rest of the paper is organized as follows. Section \ref{background} reviews background study and related work. In Section \ref{Approach-BehavDT}, we present our behavioral decision tree context-aware model. In Section \ref{Evaluation-BehavDT}, we report the evaluation results. Some key observations of our model are summarized in Section \ref{Discussion-BehavDT}. Finally, Section \ref{Conclusion} concludes this paper.

\section{Background Study and Related Work}
\label{background}
In the area of machine learning and data science, classification is well-known and most popular technique for building prediction models utilizing the given datasets \cite{han2011data}. Researchers have proposed many classification algorithms that can be used to build a predictive model. A significant number of work has been done in the area of context-aware mobile analytics based on machine learning classification techniques. Research that relies on users' multi-dimensional contextual information that are collected from individuals' smartphone data is mostly application specific. Various machine learning classification techniques are used in various context-aware applications. Among these, ZeroR is the simplest classification approach that simply predicts the majority category class \cite{witten2005data}. This classification technique is mainly used as a benchmark of all the classification algorithms. Recently, Sarker et al. \cite{sarker2019effectiveness} have used ZeroR classification technique as a benchmark in analyzing performance of various context-aware models. 

Naive Bayes \cite{john1995estimating} in one of the most popular probabilistic based classification algorithms, which can foresee the class membership probabilities \cite{han2011data}. In this classifier, the effect of a contextual attribute in the dataset on a given class is independent of those of other attributes. Pejovic et al. \cite{pejovic2014interruptme} have designed a model for the purpose of intelligently handling mobile interruptions. In their model, they have considered naive Bayes as one of the classification techniques. In addition to build a traditional predictive model, this classifier can also be used to handle the noisy instances in the datasets to make the model robust. The robustness of a predictive model is important as such noisy instances may increase the complexity of the inferred model, may increase the over-fitting problem, and thereby the overall prediction accuracy of the model may decrease. For instance, Sarker et al. \cite{sarker2019machine} have shown that how naive Bayes classifier is used to make a context-aware predictive model robust. Another classification algorithm, K-nearest neighbors \cite{aha1991instance} is also popular to build a predictive model in the area of machine learning. It is also known as instance-based learning, or lazy learning. Support vector machines \cite{keerthi2001improvements}, is another popular classification technique used widely for various predictive analytics. To do this, a hyperplane is chosen, which is a line that can distinguish the data points into different classes. A logistic function based popular classification approach is known as Logistic Regression \cite{le1992ridge}. Typically, logistic regression estimates the probabilities using that logistic function, which is also referred to as sigmoid function. In a context-aware analysis, Sarker et al. \cite{sarker2019effectiveness} have shown the effectiveness of these classification models considering multi-dimensional contexts through a range of experimental analysis.

A tree-like model is capable to build a prediction model as well as generating classification rules for making decisions. A straightforward classification approach is OneR proposed by Holte et al. \cite{holte1993very}. OneR is a simple one rule based classification algorithm. In this approach, a one-level decision tree is constructed from the training records and the rules are extracted from that tree, which are linked with frequent classes in the given dataset. Decision tree is a very well-known and most discussed technique for classification and prediction, which is used most frequently in the area of machine learning, data science, and context-aware mobile analytics as well. The first algorithm is known as ID3 decision tree \cite{quinlan1986induction} that constructs a tree by employing a top-down approach following a greedy search. To select the best attribute it uses the statistical measures entropy and information gain. Based on the ID3 algorithm, a modified and improved algorithm is proposed by Quinlan, namely C4.5 algorithm \cite{quinlan1993}, which handles the attributes more effectively. PART is a hybrid classification algorithm was proposed in \cite{frank1998generating}. In PART, a partial decision tree is constructed by using a divide and conquer approach. Like the PART algorithm, DTNB proposed by Sheng et al. \cite{sheng2005hybrid} is another hybrid classification technique for generating classification rules. It uses both the decision table and naive Bayes classifier. The produced rules can be used to predict the unseen classes for a particular condition. These classification techniques are able to model user behavioral activities based on multi-dimensional contexts \cite{sarker2019effectiveness}. In another study, Pielot et al. \cite{pielot2014didn}, have used similar classifiers in their instance message based mobile analytics to predict the attentiveness.

Among the classification techniques discussed above, decision tree \cite{quinlan1993} is one of the most popular because of having several advantages, such as easy to interpret; ability to handle multi-dimensional attributes; processing speed and simplicity in design; acceptable prediction accuracy; and the ability to generate human understandable rules \cite{wu2016decision} \cite{wu2008top}. Decision tree classification approach is frequently used in the area of context-aware systems and services. For instance, Hong et al. \cite{hong2009context} propose a context-aware model based on decision tree classifier for the purpose of providing personalized services. In their approach, they utilize context historical data to build the model. Lee et al. \cite{lee2007deploying} have proposed a decision tree based model for the purpose of deploying personalized services for the benefit of mobile phone users. In \cite{zulkernain2010mobile}, Zulkernain et al. have designed an intelligent mobile interruption management system to intelligently assist the mobile phone users in their daily activities. In their system, they use decision tree rule based technique to make their system intelligent. In \cite{sarker2019machine}, Sarker et al. have used a decision tree classifier in their machine learning based user behavior model utilizing contextual smartphone data. In addition to these approaches,  Pielot et al. \cite{pielot2014large} have designed a model to predict phone call availability. In their model, they have considered multiple decision trees like random forest classification approach to make the model more effective. Similarly, random forest classification model has also been used in Sarker et al. \cite{sarker2019AppsPred} and Sarker et al. \cite{sarker2019miim} for modeling smartphone apps usage and to build an ensemble learning based intelligent telephony model respectively. Overall, a tree like model has been used widely in the area of context-aware mobile analytics. The authors in \cite{sarker2019effectiveness} have also shown through the experimental analysis that a tree-based classification model is more effective to model and predict mobile usage behavior. However, in many context-aware cases, decision tree causes over-fitting problem and the generated classification rules mostly have low reliability \cite{sarker2019context}. Such issues may arise because of lacking generalization while making decisions in a context-aware model.

Unlike the above classification techniques and context-aware models, in this paper, we present a \textit{behavioral decision tree} machine learning classification approach that takes into account \textit{behavior-oriented generalization} and build an effective user-centric context-aware predictive model.

\section{Methodology: A BehavDT Model}
\label{Approach-BehavDT}
In this section, we present our behavioral decision tree (BehavDT) context-aware model including the behavior-oriented generalization approach by taking into account multi-dimensional contexts.

\subsection{Approach Overview}
\begin{figure*}[htbp!] 
	\centering    
	\includegraphics[width=.7\linewidth ]{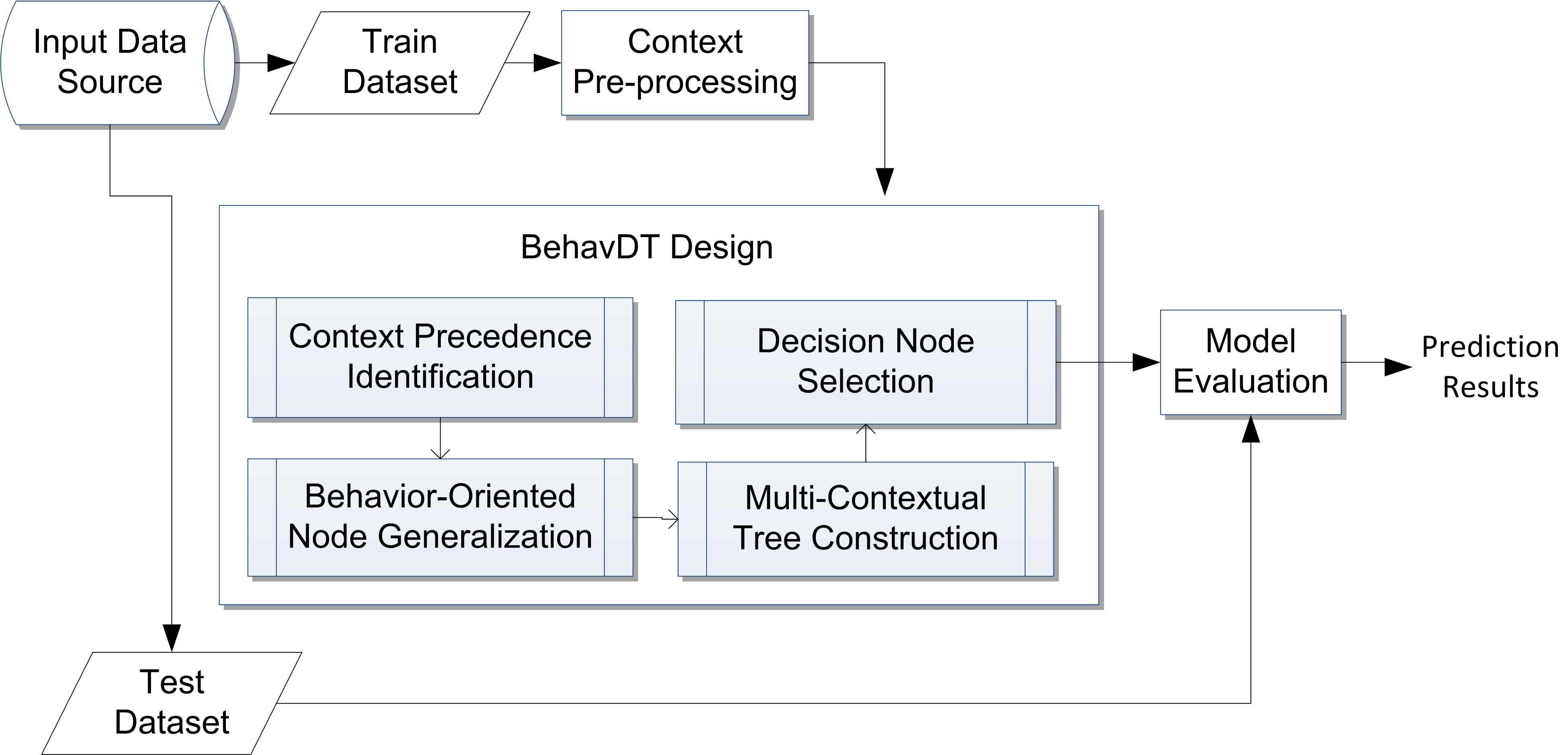}
	\caption{An overview of our BehavDT context-aware predictive model.}
	\label{fig:architecture}
\end{figure*}

Our approach accepts as input source a phone log dataset of an individual user consisting of contextual information and her mobile phone usage history. From this multi-dimensional contextual data, our model is able to predict their usage behavior by going through several processing steps. We first organize the dataset with relevant contexts such as temporal, spatial, or social, and corresponding mobile phone usage behavior of individuals. After that, we pre-process the datasets and prepare for building a decision tree classification model. Once the train dataset is ready to process, we then construct a behavioral decision tree rather than the traditional decision tree. In our BehavDT model, we take into account the precedence of contexts and perform behavior-oriented generalization while designing the tree. To create a node, the confidence value is calculated based on the dominant behavior and the associated contexts. Once the behavioral decision tree with multi-level contexts has been built, both the interior and leaf nodes are used as the decision making nodes. As a result, this model is able to capture not only the general behavior of a user but also the exceptions for the given context-aware cases. Finally, we evaluate the resultant context-aware model using the test datasets. Figure \ref{fig:architecture} shows an overview of our behavioral decision tree context-aware model.

\subsection{Context Pre-processing}
As we take smartphone raw data as input data source, shown in Figure \ref{fig:architecture}, we need to prepare and preprocess the raw dataset for designing the behavioral decision tree. In our work, we use the reality mining datasets \cite{eagle2006reality} that were collected by Massachusetts Institute of Technology, USA for their reality mining project. We first consider the contextual information such as temporal, spatial, or social contexts that might have an influence on individuals' usage behavior and organize the datasets by ignoring others metadata. We then transform the continuous contextual data into meaningful categorical values. In this process, we first discretize the continuous time-series data into time periods. For this purpose, we employ our earlier BOTS technique \cite{sarker2017individualized} that generates a number of behavior-oriented time segments according to their behavioral patterns. We also generate data-centric social context to preprocess the social relational context. While generating data-centric social context, it represents individual's one-to-one relationship using their unique contact numbers in the dataset \cite{sarker2018DataCentricSocialContext}. Calendar based social situations such as meeting, lunch break, lecture etc. could be another type of social contextual information for an individual user \cite{sarker2019calbehav}. For spatial context, we use individual's recorded location, such as home, office, market, MIT, Harvard etc. exist in the given datasets.

\subsection{Behavior-Oriented Generalization}
\begin{figure*}[htbp!]
	\centering
	\begin{subfigure}{.6\textwidth}
		\centering
		\includegraphics[width=\linewidth, height = 5cm]{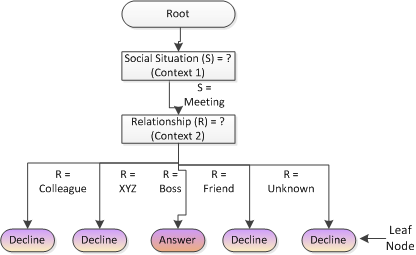}
		\captionof{figure}{Nodes in traditional decision tree.}
		\label{fig:traditioanl}
	\end{subfigure}%
	\begin{subfigure}{.3\textwidth}
		\centering
		\includegraphics[width= \linewidth, height = 5cm]{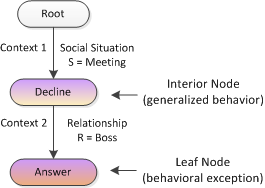}
		\captionof{figure}{Nodes based on behavior-oriented generalization in BehavDT.}
		\label{fig:generalized}
	\end{subfigure}
\caption{An example of node creation considering contextual features, where Fig \ref{fig:traditioanl} shows the nodes created in traditional decision tree \cite{quinlan1993} and Fig \ref{fig:generalized} shows the corresponding nodes created based on behavior-oriented generalization in our BehavDT model.}
\label{fig:traditional-vs-generalization}
\end{figure*}

This generalization is one of the significant features of our BehavDT context-aware model. It creates nodes according to the similar behavioral patterns of individual users for a particular confidence level preferred by them, say, 80\%. Using this generalization process, we create interior node as a generalized node rather than creating a number of single context-value specific separate nodes. Let's consider an example of phone call response behavior of an individual from several social relationships, say, friend, colleague, XYZ, boss, or unknown, when she is in a meeting. Rather than producing all separate branches for each relationship value in the tree, we first generalize these activities into one behavior class (e.g., decline) by assuming the calculated confidence satisfies user preferred threshold (say, 80\%) and create a `decline' node. Such generalized node is determined by considering the dominant characteristics that has the highest number of occurrences for that associated contexts. As there is an exceptional behavioral activities for boss's calls, (i.e., different behavior class of the generalized parent node, which is generated earlier), we create additional child node by specifying the additional context level and corresponding behavior (e.g., answer). Hence, the first created generalized node is called the interior or internal node, and the second node containing an exception with it's parent node, is called the leaf node as no further exceptions are found in this branch. Figure \ref{fig:traditional-vs-generalization} shows the generated branches of traditional decision tree and the corresponding generalization of our BehavDT model respectively for the above scenario.

The process for behavior-oriented node generalization is set out in Algorithm \ref{alg:node-generalization}. Input data includes training dataset: $DS = {X_1,X_2,...,X_n}$, which contains a set of training instances and their associated class labels, context-association $assoc$, user preferred confidence threshold $t$, and output data is the generalized node $GN$. In Algorithm \ref{alg:node-generalization}, the method identifyDominantBehavior() determines the dominant behavior for a particular $assoc$, which simply represents the most occurrences comparing with others. We define the dominant behavior $BH_{dominant}$ of a user for a particular $assoc$, which represents a particular activity that most commonly occurs among a list of behavioral activities for that $assoc$ by taking into account the relevant activity instances in the given dataset. Let  $Oc = \{Oc_1,Oc_2, ..., Oc_n\}$ be a list of activity occurrences in percentage (\%) and $n$ is the number of behavior classes for a particular $assoc$, then $BH_{dominant} = MAX({Oc_1,Oc_2,...,Oc_n})$ represents the dominant behavior for that $assoc$. Another method calculateConfidence() calculates the confidence value by determining the ratio of the dominant behavior and corresponding $assoc$. This confidence measures the strength of a node whether it is reliable to make a decision or not for a particular $assoc$. Finally, based on this confidence value and dominant behavior this algorithm creates the generalized node $GN$ for that particular $assoc$.

\begin{algorithm}[htbp!]
	\caption{Behavior-Oriented Generalization}
	\label{alg:node-generalization}
	\SetKwInOut{Data}{Data}
	\SetAlgoLined
	\Data{Mobile phone dataset $DS = {X_1,X_2,...,X_n}$ // each instance $X_i$ contains a number of nominal context-values, context-association $assoc$, user preferred confidence threshold $t$}
	
	\KwResult{generalized node: $GN$}
	
	\BlankLine
	
	\underline{Procedure NodeGeneralization} $(DS, assoc, t)$\;
	
	//create subset containing the particular context-association \\
	$DS_{sub} \leftarrow$ subset of $DS$ that contains $assoc$ \\
	
	\If{$DS_{sub} \neq \phi$}
	{
		//identify the dominant behavior that represents the most occurrences. \\
		$BH_{dominant} \leftarrow identifyDominantBehavior(DS_{sub})$ \\
		//calculate the confidence for that identified dominant behavior \\
		$t_{conf} \leftarrow calculateConfidence(DS_{sub}, BH_{dominant})$ \\
		\If{$t_{conf}$ $>=$ t}
		{
			//create new generalize node in the tree \\
			$GN \leftarrow createNode ()$\\
			// assign behavior class to the new node \\
			$GN_{behavior} \leftarrow BC_{dominant} $\\
		}	
		
		return $GN$;	
	}
	
\end{algorithm}

\subsection{Decision Nodes in BehavDT}
A BehavDT is a tree structure that consists of different nodes such as a root node, interior and/or leaf nodes, and associated branches. Each branch in BehavDT denotes a contextual test on a context-value (e.g., is interpersonal social relationship value `boss'?), and each node (interior or leaf) denotes the behavioral outcome of that test which is represented by a behavior activity class label (e.g., answer). The topmost node in the tree is called the root node which is an empty node. The all other nodes are created according to the test and corresponding behavioral outcome. 

\begin{figure*}[htbp!] 
	\centering    
	\includegraphics[width=.6\linewidth]{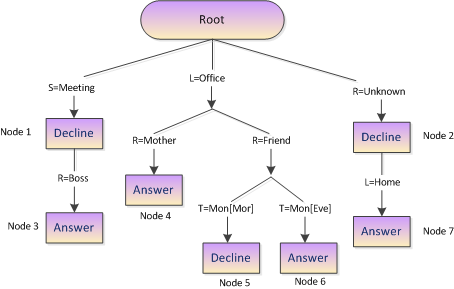}
	\caption{An example of a behavioral decision tree including interior and leaf decision nodes based on contexts.}
	\label{fig:tree}
\end{figure*}

In order to determine all the decision nodes in BehavDT, we first construct a complete BehavDT based on multi-dimensional contexts. To build a complete BehavDT, we follow a top-down approach, starting from a root node. The tree is partitioned into behavioral activity classes distinguished by the values of the most discriminant context determined by calculating the precedence of contexts. Such precedence of contexts is determined by calculating the entropy and information gain. Formally, entropy is defined as below \cite{witten2005data}:

\begin{equation}
\label{entropy}
$$H(S)= -\sum_{x \in X} p(x)log_2p(x)$$
\end{equation}

Where, $S$ is the current data set for which entropy is being calculated, $X$ represents a set of classes in $S$, $p(x)$ is the proportion of the number of elements in class $x$ to the number of elements in set $S$. The formal definition of information gain is as below \cite{witten2005data}:

\begin{equation}
\label{gain}
$$IG(A,S)= H(S)-\sum_{t \in T} p(t)H(t)$$
\end{equation}

Where, $H(S)$ is the entropy of set $S$, $T$ represents the subsets created from splitting set $S$ by attribute $A$ such that $S=\cup_{t \in T} t$, $p(t)$ is the proportion of the number of elements in $t$ to the number of elements in set $S$, $H(t)$ is the entropy of subset $t$. \\

We dynamically use this precedence of contexts in each node to effectively choose the next significant context rather than taking into account same context in each level. Once the root node of the BehavDT has been determined, we then create the child nodes for the associated contexts. To create nodes, we perform behavior oriented generalization that has been discussed earlier. The algorithm continues recursively by adding new subtrees to each context branch and terminates when all the instances in the reduced training set belong to the same behavior class, or getting the context list empty. The overall step by step procedure for learning the behavioral decision tree as follows:

\begin{enumerate}
	\item The approach takes input dataset: $DS = {X_1,X_2,...,X_n}$, where each instance $X_i$ contains a number of contextual attributes and corresponding behavior class $BH$. 
	\item Create a root node and assign all training instances to the root of the tree.
	\item For each contextual attribute:
	  \begin{itemize}
	  	\item Split the training dataset into subsets, in such a way that each subset contains data with the same value of the attribute.
	  	\item Compute entropy and information gain from the subsets using formulas defined in Equation \ref{entropy}, and Equation \ref{gain}.
	  \end{itemize}
  \item Determine the best contextual attribute based on information gain value and set this feature to be the splitting criterion at the current node.
  \item Partition all instances according to attribute value of the best feature.
  \item Denote each partition as a child node of the current node.
  \item For each child node:
  	  \begin{itemize}
  	  	\item If the child node is ``pure'', i.e., all the instances contain same behavior class, label it as a leaf node with that class and return. 
  	  	\item If not then create a behavior-oriented generalized node using Algorithm \ref{alg:node-generalization} and set the generalized node as the current node. If no generalized node found create all the context-specific nodes.
  	  \end{itemize}
  \item Recursively continue this procedure for all branches until generating the complete behavioral decision tree according to the relevant contexts.
\end{enumerate}

The final result of our BehavDT is a multi-level tree with a number of interior and leaf nodes, and their associated contexts. Figure \ref{fig:tree} shows an example of a behavioral decision tree containing multi-dimensional contexts and corresponding phone call behavior classes for a particular confidence preference 80\%. In Figure \ref{fig:tree}, $S$ represents social situation context, such as meeting; $L$ represents spatial context location, such as office, home; $R$ represents social relational context such as boss, mother, friend, unknown; and $T$ represents temporal context such as Monday morning, Monday evening etc. Once the tree has been built, both the interior and leaf nodes are used as the decision making nodes in our BehavDT model. Hence, Table \ref{node-descriptions} gives an overview of the generated decision nodes in the tree shown in Figure \ref{fig:tree}.

\begin{table*}[htbp!]
	\centering
	\caption{An overview of the decision nodes in the tree shown in Figure \ref{fig:tree}}
	\label{node-descriptions}
	\begin{tabular}{|c|c|c|c|} 
		\hline
		\bf Node No & \bf Node Type & \bf Behavior & \bf Associated Contexts \\  
		\hline
		Node 01 & Interior & Decline & Meeting \\ 
		\hline
		Node 02 & Interior & Decline & Unknown \\ 
		\hline
		Node 03 & Leaf & Answer & Meeting, Boss \\ 
		\hline
		Node 04 & Leaf & Answer & Office, Mother \\ 
		\hline
		Node 05 & Leaf & Decline & Office, Friend, Mon[Mor] \\ 
		\hline
		Node 06 & Leaf & Answer & Office, Friend, Mon[Eve] \\ 
		\hline
		Node 07 & Leaf & Answer & Unknown, Home \\ 
		\hline
	\end{tabular}
\end{table*}

\section{Evaluation and Experimental Results}
\label{Evaluation-BehavDT}	 
To evaluate our proposed BehavDT context-aware model, we have conducted experiments on individual users' smartphone datasets. We have conducted experiments on ten phone log datasets collected by Massachusetts Institute of Technology (MIT) over the period of nine months for their Reality Mining project~\cite{eagle2006reality}. These datasets consist of multi-dimensional contexts like temporal, spatial, or social contexts, and corresponding phone usage behavior of the users. These are represented as U01, U02, ..., U10 for ten individual mobile phone users respectively for experimental purpose. In the following, we discuss our several findings in experiments.

\subsection{Evaluation Metric}
To evaluate our BehavDT context-aware model, we utilize a 10-fold cross validation technique on each dataset. For this, we randomly divide each dataset into ten sub parts. We then build our model using nine parts and use the remaining part as test dataset, and measure the effectiveness in terms of prediction accuracy. For this, we compare the predicted response with the actual response (i.e., the ground truth) and compute precision, recall, and overall accuracy. If TP denotes true positives, FP denotes false positives, TN denotes true negative, and FN denotes false negatives, then these evaluation metrics are defined formally as below \cite{han2011data}:

\begin{equation}
Precision = \frac{TP}{TP + FP}
\end{equation}

\begin{equation}
Recall = \frac{TP}{TP + FN}
\end{equation}

\begin{equation}
Accuracy = \frac{TP + TN}{TP + TN + FP + FN}
\end{equation}

\subsection{Effect of Generalization on Decision Nodes}
\begin{figure}[htbp!]
	\centering
	\includegraphics[width=\linewidth, keepaspectratio]{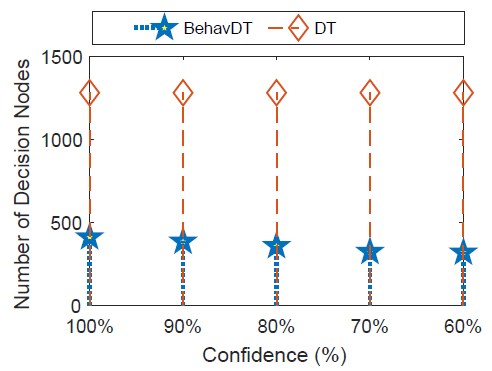}
	\caption{The effect of generalization on the number of decision nodes in our BehavDT model.}
	\label{fig:comparison}
\end{figure}

To show the effect of generalization on the number of decision nodes, we have shown the results in Figure \ref{fig:comparison} for both the traditional decision tree and the proposed BehavDT context-aware model. To show the fare comparison, we use the same datasets mentioned above for each model and show the average results. If we observe Figure \ref{fig:comparison}, we see that our BehavDT model produces comparatively less number of decision nodes for different confidence levels, 100\%, 90\%,..., 60\%, preferred by individuals. The reason is that our BehavDT model performs behavior-oriented generalization. On the other hand, traditional decision tree model does not take into account such generalization. It takes into account all the leaf position nodes as decision nodes and makes the number of decision nodes unnecessarily large. The behavior-oriented generalization in BehavDT model subsumes several context-specific nodes into one generalized node according to the similar behavioral patterns of an individual user. As a result, our BehavDT model significantly reduces the total number of decision nodes effectively while comparing with traditional decision tree model for a particular confidence level, shown in Figure \ref{fig:comparison}.

\subsection{Effect of Confidence on Precision and Recall}
\begin{figure*}[htbp!]
	\centering
	\begin{minipage}{.45\textwidth}
		\centering
		\includegraphics[width=\linewidth, height = 4cm]{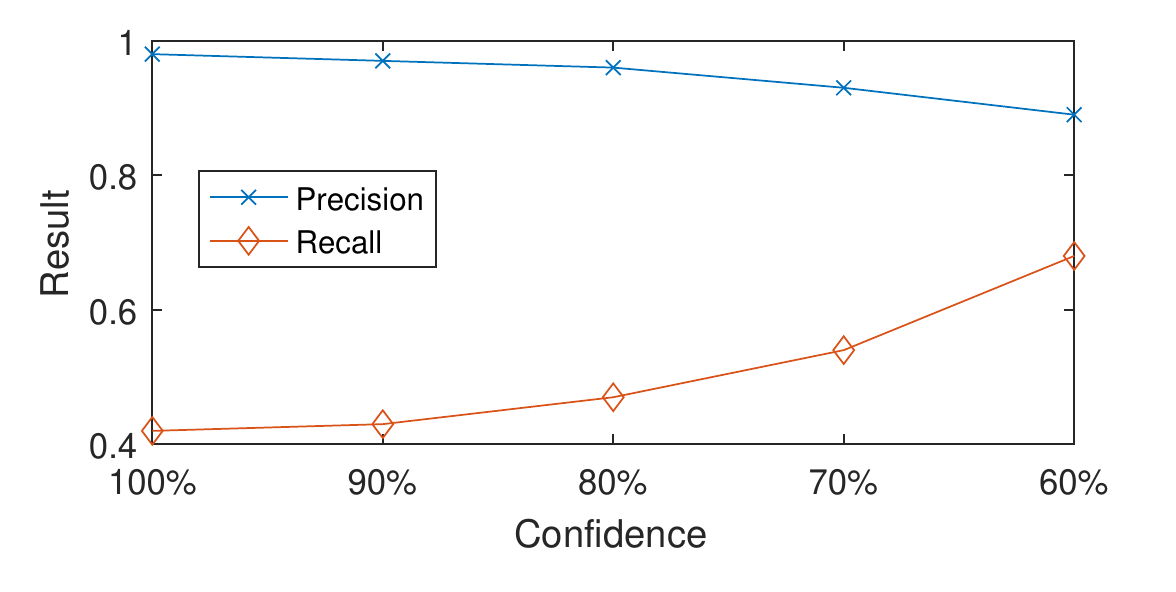}
		\captionof{figure}{Prediction results in terms of precision and recall using different confidence levels for user U01.}
		\label{fig:precision-recall-RM-10}
	\end{minipage}%
    \hspace{0.05\textwidth}
	\begin{minipage}{.45\textwidth}
		\centering
		\includegraphics[width= \linewidth, height = 4cm]{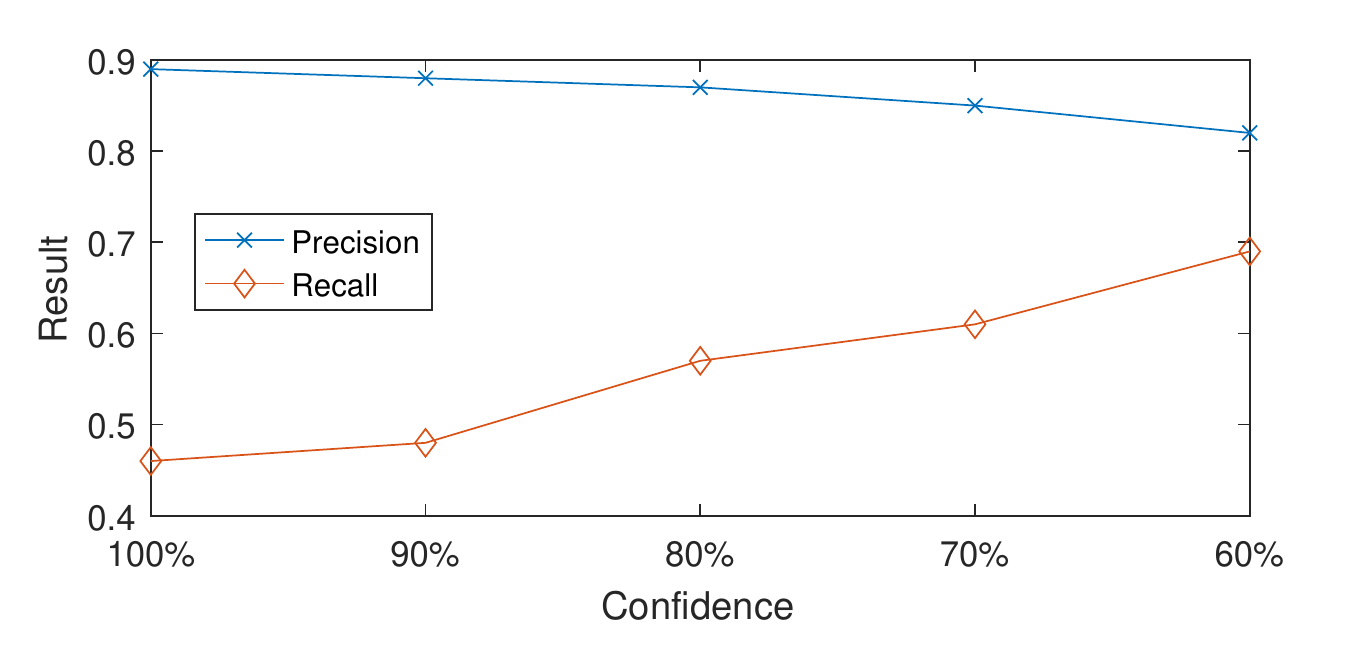}
		\captionof{figure}{Prediction results in terms of precision and recall using different confidence levels for user U02.}
		\label{fig:precision-recall-RM-15}
	\end{minipage}
\end{figure*}

In this experiment, we show the effect of confidence on the prediction results for different users utilizing their datasets. For this, we first illustrate the prediction results in terms of precision and recall by varying the conference values. We take into account the confidence values from 100\% (maximum) below to 60\% (lowest). Since confidence is associated to accuracy level, we are not interested to take into account below 60\% as confidence preference in this experiment. Figure \ref{fig:precision-recall-RM-10} and Figure \ref{fig:precision-recall-RM-15} show the relationship between precision and recall for different confidence thresholds for two users utilizing their phone datasets. As our model is individual user-centric, we illustrate the results for individual users.

Typically, higher precision results in a lower recall and vice-versa. In our BehavDT model, we use a confidence preference for making reliable prediction in a particular context, to decide whether a prediction result is significant or not for the user. If we observe Figure \ref{fig:precision-recall-RM-10} and Figure \ref{fig:precision-recall-RM-15}, we see that recall increases with the decrease of confidence levels. The main reason for increasing recall with the low confidence threshold is that a large number of context-specific decision nodes satisfy this low confidence threshold and are subsumed into the generalized one. Thus, the generalized node is able to make predictions for similar context-aware test cases considering low confidence. With the increasing of the confidence level, the number of context-specific decision nodes satisfying this threshold are decreasing, and as a result recall is also decreasing. On the other hand, precision increases with the increase of confidence level. If the confidence threshold is low, large number of incorrect predictions are generated. With the increasing of the confidence level, the incorrect predictions are decreasing and the resulting precision is increasing. Using a higher confidence threshold results in higher precision but lower recall, and using a lower confidence threshold results in lower precision but higher recall. By default, we use 80\% confidence preference to select decision nodes for building the prediction model assuming the common preference of all users. Further, we allow users to configure the recall-precision trade off based on their individual preferences.

\subsection{Effectiveness Comparison}
In this experiment, we show the effectiveness of our BehavDT model in terms of prediction accuracy, comparing it with existing popular classification approaches. Among the baseline approaches, decision tree (DT) is used mostly to build a context-aware model for the purpose of providing various mobile services \cite{hong2009context} \cite{lee2007deploying} \cite{zulkernain2010mobile} \cite{sarker2019machine}, that are discussed in Section \ref{background}. In addition to decision tree, ZeroR, naive Bayes (NB), support vector machines (SVM), logistic regression (LR), k-nearest neighbor (KNN) are also used in our comparison purposes. The details of these classification based context-aware models are discussed in Section \ref{background}. Although, decision tree model is highly relevant to our BehavDT model, we compare with all these baseline approaches in our effectiveness analysis. 

\begin{figure*}[htbp!]
	\centering
	\begin{minipage}{.45\textwidth}
		\centering
		\includegraphics[width= \linewidth, height = 4.5cm]{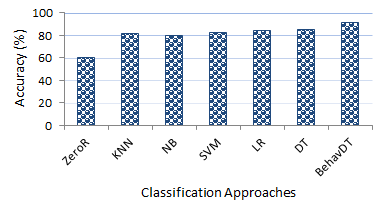}
		\captionof{figure}{Comparison of prediction results in terms of accuracy (\%) of different classification models for user U01.}
		\label{fig:accuracy-comparison-RM-10}
	\end{minipage}%
    \hspace{0.05\textwidth}
	\begin{minipage}{.45\textwidth}
		\centering
		\includegraphics[width= \linewidth, height = 4.5cm]{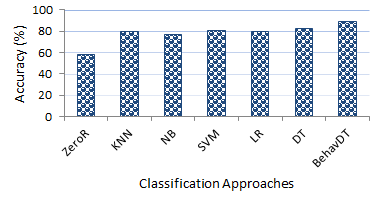}
		\captionof{figure}{Comparison of prediction results in terms of accuracy (\%) of different classification models for user U02.}
		\label{fig:accuracy-comparison-RM-15}
	\end{minipage}
\end{figure*}

In our effectiveness analysis, we have shown the comparison of prediction results of different classification models mentioned above for different users. Figure \ref{fig:accuracy-comparison-RM-10} and Figure \ref{fig:accuracy-comparison-RM-15} show the prediction results in terms of accuracy (\%) for various context-aware classification models for user U01 and U02 respectively. For the purpose of fare comparison, we utilize the same datasets in both training and testing sets for all these approaches. From Figure \ref{fig:accuracy-comparison-RM-10}, and Figure \ref{fig:accuracy-comparison-RM-15}, we find that our BehavDT context-aware model consistently outperforms traditional machine learning approaches for predicting individual user's behavior in multi-dimensional contexts that we have considered in this study. The reason is that we take into account behavior oriented-generalization in our BehavDT model and use the generalized nodes as decision nodes. In addition, we also take into account context-specific nodes as decision nodes for relevant exceptions. As a result, our BehavDT model not only determines the significant decision nodes but also able to capture both the general behavior of an individual user and the specific exceptions as well. Such capability of our BehavDT model improves the prediction accuracy for different users shown in Figure \ref{fig:accuracy-comparison-RM-10} and Figure \ref{fig:accuracy-comparison-RM-15}, and makes the resultant machine learning based user-centric context-aware predictive model more effective.

\section{Discussion: Traditional DT vs BehavDT}
\label{Discussion-BehavDT}
Overall, our BehavDT model is applicable to effectively predict individual mobile phone users behavior in their various daily life situations. Compared to the existing classification approaches, the prediction accuracy (\%) has been improved when our BehavDT model is used, as shown in Figures \ref{fig:accuracy-comparison-RM-10}, \ref{fig:accuracy-comparison-RM-15}. Among these existing classifiers, decision tree \cite{quinlan1993} is the most relevant technique with our BehavDT in terms of the tree-like structure. In the following, we highlight some key differences between traditional decision tree \cite{quinlan1993} and our proposed BehavDT model for predicting user diverse behaviors in multi-dimensional contexts.

\begin{itemize}
	\item \textit{Individualized preference-oriented:} The traditional decision tree does not provide flexibility to set user preferences that lead to rigid decision making. Such rigid decisions might not be suitable to model individuals' behavioral activities, because of their day-to-day variations in their daily life activities. On the other hand, our BehavDT model specifically designed to build user-centric context-aware predictive model that takes into account user preferences. As the preferences may vary from user-to-user in the real world, our BehavDT model dynamically adapts such variations while designing the tree.
	
	\item \textit{Behavior-oriented generalization:} The traditional decision tree does not consider user behavior-oriented generalization. As a result, it produces a large number of leaf position nodes that may cause over-fitting problem of the tree. On the other hand, while designing our BehavDT, we have performed generalization according to individual's similar behavioral patterns for a particular confidence level. This generalization can play a role not only to capture individuals' general behaviors but also to minimize the over-fitting problem by creating the generalized nodes.
	
	\item \textit{Tree structure:} The traditional decision tree typically generates leaf nodes containing the class values. On the other hand, in our BehavDT model, each branch denotes a test on a context value, and each node either interior or leaf may denote the outcome containing the behavior class for that test. Thus, unlike the traditional decision tree, a number of interior nodes containing the class values may exist in our BehavDT model.
	
	\item \textit{Decision making node:} In the traditional decision tree, all the leaf position nodes are typically considered as decision nodes. As a result, in many cases such decision nodes become low reliable, i.e., less performing in a user-centric context-aware predictive model. On the other hand, we do not depend on node's position for choosing the effective decision nodes. Both interior and leaf nodes generated in the tree, are used as decision nodes in our BehavDT context-aware model.
	
	\item \textit{Effectiveness:} The effectiveness of the context-aware model designed by the traditional decision tree might be low in terms of prediction accuracy, when comparing with our model. On the other-hand, our BehavDT context-aware model is more effective to predict user diverse behaviors in their various day-to-day situations.
\end{itemize}

The above key differences make our BehavDT model more effective to build a user-centric context aware predictive model over the traditional decision tree model. As data-driven RecencyMiner \cite{sarker2019recencyminer} can play a significant role in modeling user behavior, our BehavDT model can also be used for further analysis in terms of recency in relevant context-aware problems and services.

\section{Conclusion}
\label{Conclusion}
In this paper, we have presented a behavioral decision tree, BehavDT machine learning approach to build user-centric context-aware predictive model. Our BehavDT model predicts smartphone user behavioral activities based on multi-dimensional contexts, such as temporal, spatial, or social contexts. In our BehavDT model, we have incorporated behavior-oriented generalization while designing the decision tree. We have taken into account both the interior and leaf nodes as the decision making nodes, to effectively capture user diverse behaviors for their various day-to-day context-aware situations. We have conducted experiments on individuals' real mobile phone datasets containing their behavioral activities and corresponding contextual information. The experimental results have shown that our BehavDT model not only determines the significant decision nodes but also able to capture both the general behavior of an individual user and the specific exceptions as well. Thus, BehavDT model have shown the effectiveness for building user-centric context-aware predictive model. Although, we use phone call behavior example to make understand this BehavDT model, it can also be applicable to other user-centric application domains, where user diverse behavioral activities are impacted by the surrounding contexts. As a future work, we plan to asses the usability of this machine learning context-aware model in the application level.

\bibliographystyle{plain}
\bibliography{MobileDataAnalytics}

\end{document}